\documentclass[letterpaper, 10 pt, conference]{ieeeconf}

\IEEEoverridecommandlockouts                              

\usepackage[
top    = 54pt,
bottom = 58pt,
left   = 54pt,
right  = 54pt]{geometry}
\usepackage{booktabs}
\usepackage{caption}
\usepackage{subcaption}
\usepackage{graphics}
\usepackage{graphicx}
\usepackage{amsmath}
\usepackage{hyperref}
\usepackage{color, soul}
\usepackage{float}

\renewcommand\hl[1]{#1} 

\newcommand{\fig}[1]{Fig.~\ref{#1}}
\newcommand{\tbl}[1]{Table~\ref{#1}}

\newcommand{\sect}[1]{Sec.~\ref{#1}}
\newcommand{\etal}[0]{{\em et al.~}}
\newcommand{\eg}[0]{{\em e.g.,~}}

\title{\Large \bf Multi-modal Transfer Learning for Grasping Transparent and Specular Objects}

\author{Thomas Weng$^{1}$, Amith Pallankize$^{2}$, Yimin Tang$^{3}$, Oliver Kroemer$^{1}$, and David Held$^{1}$
\thanks{
This work was supported by the National Science Foundation Smart and Autonomous Systems Program (IIS-1849154), the Sony Corporation, the Office of Naval Research (N00014-18-1-2775), the NSF Graduate Research Fellowship Program (DGE-1745016), the Efort Intelligent Equipment Company, and ShanghaiTech University. Any opinions, findings, and conclusions or recommendations expressed in this material are those of the authors and do not necessarily reflect the views of the ONR and NSF.} 
\thanks{$^{1}$Thomas Weng, Oliver Kroemer, and David Held are with the Robotics Institute, Carnegie Mellon University, Pittsburgh, PA
        {\tt\footnotesize \{tweng, okoremer, dheld\}@andrew.cmu.edu}}%
\thanks{$^{2} $Amith Pallankize is with Microsoft Corporation, Hyderabad, India.
        {\tt\footnotesize ampallan@microsoft.com}}%
\thanks{$^{3} $Yimin Tang is with ShanghaiTech University, Shanghai, China.
        {\tt\footnotesize tangym@shanghaitech.edu.cn}}%
}

\begin{document}


\maketitle

\thispagestyle{empty}
\pagestyle{empty}

\begin{abstract}
State-of-the-art object grasping methods rely on depth sensing to plan robust grasps, but commercially available depth sensors fail to detect transparent and specular objects.
To improve grasping performance on such objects, we introduce a method for learning a multi-modal perception model by bootstrapping from an existing uni-modal model. 
\hl{This transfer learning approach requires only a pre-existing uni-modal grasping model and paired multi-modal image data for training, foregoing the need for ground-truth grasp success labels nor real grasp attempts. Our experiments demonstrate that our approach is able to reliably grasp transparent and reflective objects.
Video and supplementary material are available at} \href{https://sites.google.com/view/transparent-specular-grasping}{\hl{https://sites.google.com/view/transparent-specular-grasping}}.
\end{abstract}


\section{Introduction}
\label{sec:intro}

Robotic grasping is a key prerequisite for a variety of tasks involving robot manipulation.
Robust object grasping would enable a wide range of applications in both industrial and natural human environments. 
The challenge with grasping is that many factors influence the effectiveness of a grasp, such as gripper and object geometries, object mass distribution and friction, and environmental conditions like illumination.

\begin{figure}[t]
    \centering
    \begin{subfigure}[b]{0.455\columnwidth}
        \centering
        \includegraphics[width=0.99\columnwidth, trim=110 0 110 0, clip]{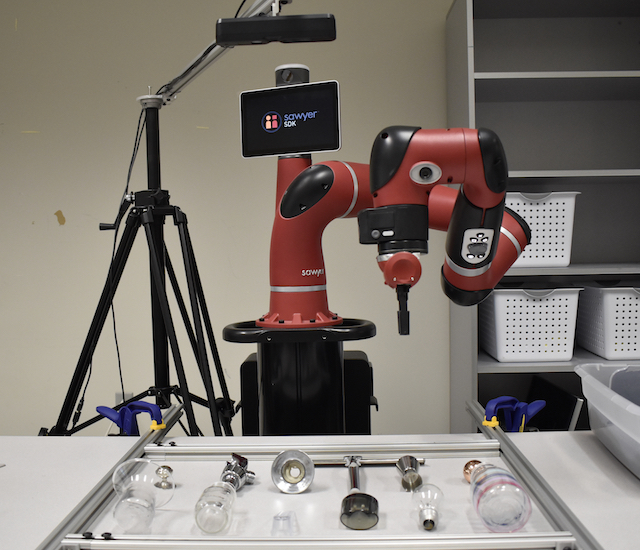}
    \end{subfigure}%
    \begin{subfigure}[b]{0.54\columnwidth}
        \centering
        \includegraphics[width=0.99\columnwidth, trim=-4 -8 0 0 0, clip]{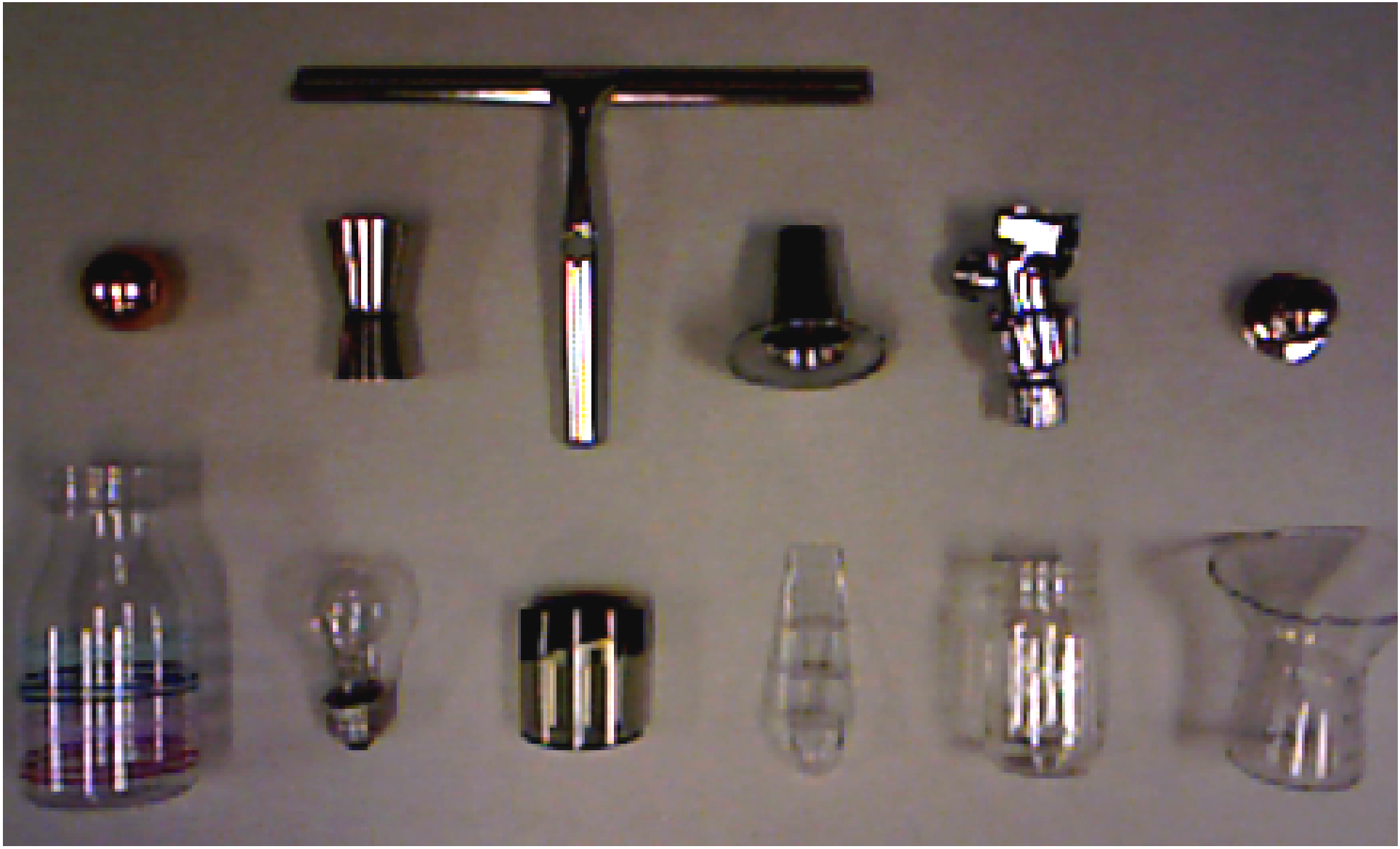}
        \includegraphics[width=0.99\columnwidth, trim=-5 0 0 0]{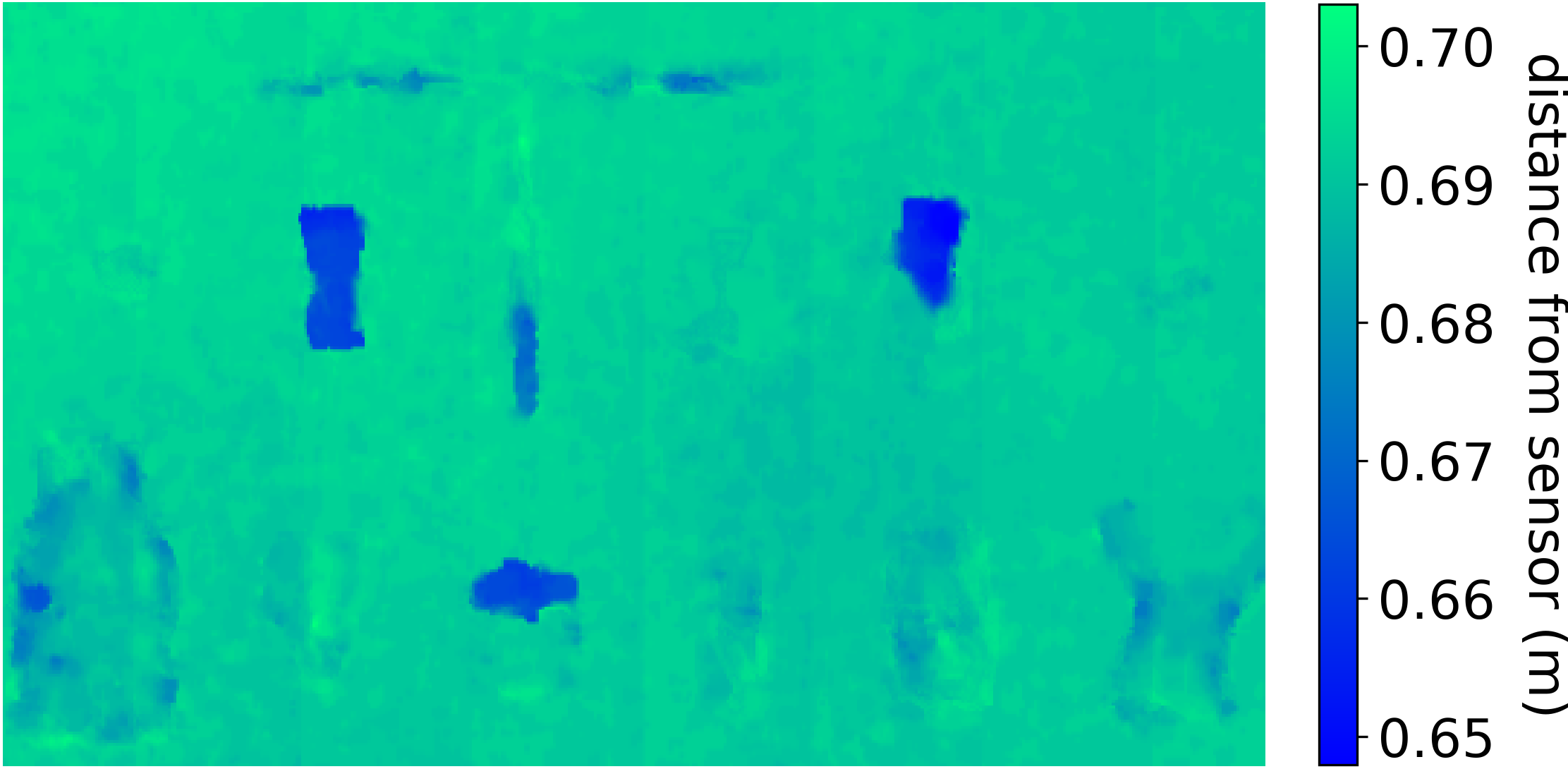}
    \end{subfigure}
    \caption{Transparent and specular objects provide poor depth readings with conventional depth sensors, posing a challenge for depth-based grasping techniques. (left) Robot workspace with fixed overhead sensor for grasping. (top right) Color image of scene from overhead sensor. (bottom right) Depth image of scene showing that most values in depth image are close to the table.}
    \label{fig:intro}
    \vspace{-1em}
\end{figure}

Most state-of-the-art grasping methods rely on depth input from structured light or time-of-flight sensors to determine the best grasp for an object~\cite{satish2019policy, morrison2018closing, gualtieri2016high}.
Under normal operation, such devices emit light patterns onto a scene and use a receiver to construct depth based on changes in the returned pattern. 
However, such depth sensors fail to detect objects that are transparent, specular, refractive, or have low surface albedo~\cite{ihrke2010transparent}, causing depth-based grasp prediction methods to fail.
These failures can take the form of both missing depth readings, as is the case with specular objects that deflect structured light patterns, and incorrect depth values, which occur when the emitted light passes through transparent objects (see \fig{fig:intro}). 

Transparent and specular objects are common in a range of environments, such as in manufacturing facilities, retail spaces, and homes. 
Under certain lighting conditions and object properties, even seemingly opaque objects can exhibit sensor noise similar to transparency and specularity.
The ubiquity of objects with these challenging properties requires us to design methods capable of bridging the sensory gap so that robots can robustly grasp a diverse set of objects. 

Our contribution in this work is a method for learning to grasp transparent and specular objects that leverages existing depth-based models.
Transparent and specular objects are more identifiable in RGB space, where transparencies and specularities produce changes in coloration, rather than the inaccurate or missing values that occur in depth space.
Therefore, we make use of both color and depth modalities in our approach.
We first train a color-based grasp prediction model from a depth-based one using \textit{supervision transfer}~\cite{hoffman2016cross}, a technique for transferring a learned representation from one modality to another. 
\hl{This transfer technique only requires paired RGB-D images and an existing depth-based grasping method from which to transfer; our method does not require robot grasp attempts nor human annotations}. 

We conduct real robot grasping experiments on both isolated objects and clutter to show that (1) the RGB-only network produces better grasp candidates for transparent and specular objects, compared to the depth-only network that it was trained from, and (2) the RGB-only network is complementary to the original depth model, such that combining the outputs of both models results in the best overall grasping performance on all three object types. 
We conduct additional experiments to demonstrate the robustness of our method against slight variations in illuminance, and we discuss failure cases as part of our analysis.

\section{Related Work}
\label{sec:related}

\subsection{Sensing Transparent and Specular Objects}

Sensing transparent and specular objects is a well-studied challenge in the computer vision community. 
Ihrke \etal\cite{ihrke2010transparent} provide a survey of recent approaches to transparent and specular object reconstruction.
Curless \etal\cite{curless1995better} perform space-time analysis on structured light sensing to achieve better triangulation on transparent objects.
Structured light sensing can also be paired with additional equipment like polarization lenses, light fields, or immersion in fluorescent or refractive liquids to detect transparent objects.
While structured light sensing is the closest to commercial sensing, the survey also presents methods that improve on multi-view stereo matching to detect transparent and specular objects.

Light field photography for depth reconstruction is another direction for detecting specular and transparent objects~\cite{6619203,wetzstein2011hand}. 
Light field photography has been used in robotics by 
Oberlin \etal\cite{oberlintime} applied light field photography to robot manipulation tasks like grasping non-Lambertian objects under running water.
However, this method requires capturing a dense set of images in a 3D volume over the scene of interest at both training and test time to construct suitable synthetic images for grasping.
In comparison, our proposed method requires a single, static RGB-D sensor, resulting in faster and simpler training and deployment.

Commercial RGB-D sensors (\eg~Intel RealSense, Microsoft Kinect, PrimeSense) use structured-light or time-of-flight techniques to estimate depth.
These techniques fail on transparent and specular surfaces, either allowing light emitted by the sensor to pass through or scattering it by reflection.
IR stereo and cross-modal stereo techniques have been used to improve depth reconstruction, but the reconstruction quality is still not comparable to that of Lambertian, or diffusely reflective, objects~\cite{alhwarin2014ir, chiu2011improving, mahler2018guest}. 
Lysenkov \etal\cite{lysenkov2013recognition, lysenkov2013pose} painted over transparent objects to create a dataset of paired transparent and opaque objects, but this approach scales poorly for objects with arbitrary geometries and material properties.
Our proposed method is able to use conventional RGB-D sensors without hardware and environmental modifications by combining depth and color information. 

\subsection{Grasp Synthesis}

Grasp synthesis refers to the problem of finding a stable robotic grasp for a given object and is a longstanding research problem in robotics. 
Approaches to grasp synthesis can be classified into analytic and empirical methods; see Bohg \etal\cite{bohg2013data} for a survey.
Analytic approaches use physics-based contact models to compute force closure on an object, using the shape and estimated pose of the target object~\cite{miller2003automatic, ten2017grasp, watkins2018multi}, but work poorly in the real world due to noisy sensing, simplified assumptions of contact physics, and difficulty in placing contact points accurately.

Empirical approaches, on the other hand, learn to predict the quality of grasp candidates from data on a diverse set of objects, images, and grasp attempts collected through human labeling~\cite{saxena2008robotic, jiang2011efficient, doi:10.1177/0278364914549607, 7139361}, self-supervision~\cite{pinto2016supersizing, levine2018learning}, or simulated data~\cite{depierre2018jacquard, doi:10.1177/0278364919859066, gualtieri2016high, mahler2017dex, satish2019policy}. 
Saxena \etal\cite{saxena2008robotic} trained a classifier on human-labeled RGB images to predict grasp points, triangulated the points on stereo RGB images, and demonstrated successful grasps on a limited set of household objects, including some transparent and specular objects.
However, the predicted grasp points for transparent and specular objects were limited to grasps on points where stereo triangulation was successful. 
The Cornell Grasping Dataset~\cite{jiang2011efficient}, consisting of 1k RGB-D images of objects and human-labeled grasps parameterized as an oriented bounding box, has been used to train many deep learning-based grasping methods~\cite{doi:10.1177/0278364914549607, doi:10.1177/0278364919859066, 7139361}. 
Self-supervised methods such as those by Pinto and Gupta~\cite{pinto2016supersizing} or Levine \etal\cite{levine2018learning} forego the need for human labels by training a robot to grasp directly from real grasp attempts, but these methods require tens of thousands of attempts to converge.

Recently, approaches trained on data gathered in simulation have demonstrated state-of-the-art performance.
The Jacquard dataset by Amaury \etal\cite{depierre2018jacquard} uses a grasp specification similar to the Cornell Grasping Dataset, contains simulated objects and grasp attempts, and has been successfully used for training by Morrison \etal's GG-CNN~\cite{doi:10.1177/0278364919859066}. 
Mahler \etal\cite{mahler2017dex} developed GQCNN, which was trained on a dataset of simulated grasps generated using analytic model, representing a hybrid empirical and analytic approach.  

As we will show, these depth-only grasping approaches fail on transparent and reflective objects. 
Note that GG-CNN could be modified to incorporate RGB images, which could potentially be used to grasp transparent and specular objects after training on simulated images (such as those in the Jacquard dataset~\cite{depierre2018jacquard}); however, such performance has not been demonstrated; this method has only been demonstrated for depth-based grasping of opaque objects.
In this work, we build upon the fully convolutional version of GQCNN (FC-GQCNN) proposed by Satish \etal\cite{satish2019policy}, but our method is agnostic to the specific network architecture used.
Our method does not require any real-world grasps or labeled data but instead relies on supervision transfer from \hl{a pre-trained depth network to obtain a multi-modal grasping method. The pre-trained depth network also may not require real-world grasps or human labels; for example, FC-GQCNN is trained entirely on simulated grasps.}

\subsection{Cross-modal Transfer Learning}
Supervision transfer has been explored in the past for tasks such as image classification and object detection~\cite{gupta2016cross, hoffman2016cross, li2018cross}.  These approaches are typically used to transfer image-based networks trained on ImageNet~\cite{deng2009imagenet} to depth-based or RGB-D based classification or detection networks.  To our knowledge, such approaches have not been used previously in the context of multi-modal grasping.  We show that such an approach can lead to greatly improved performance for grasping transparent and reflective objects, and can even improve performance on some opaque objects.

\section{Approach}
\label{sec:method}
Here we describe our approach for supervision transfer, which enables us to transfer a grasping method trained in one modality $\mathcal{M}_d$ to also incorporate an additional modality $\mathcal{M}_s$ without needing \hl{any additional real grasp attempts, simulation, nor human-labeled data (other than the data used to train the initial uni-modal grasping method, which in our case is only simulated rendered depth data}~\cite{satish2019policy}).

\subsection{Problem Statement}
We assume that we initially have a grasping method that takes input from a given modality $\mathcal{M}_d$, such as depth.  
Specifically, we assume that we have a grasping method that, given a candidate grasp $q$ and an image $I_d$ of modality $\mathcal{M}_d$ (\eg a depth image), outputs a grasp score $G(q, I_d)$. \hl{We wish to transfer this scoring method to a new input modality $\mathcal{M}_s$ (\eg RGB). Ideally, this new modality $\mathcal{M}_s$ will allow our grasping method to succeed in grasping certain types of objects (e.g. transparent and specular) where the previous modality, $\mathcal{M}_d$, failed. In later sections, we will discuss combining these modalities to create more robust grasping methods}.

\hl{We assume access to} a dataset of image pairs $(I_d, I_s)$, where each pair consists of one image from each modality.  We assume that \hl{each pair of images was} taken at approximately the same time and thus represent images of the same scene under the two modalities $\mathcal{M}_d$ and $\mathcal{M}_s$. \hl{Paired images for RGB and depth modalities can be captured using commercially available} RGB-D sensors (\eg Intel RealSense, Microsoft Kinect, PrimeSense).

Note that these paired images can be collected without needing to perform any grasp attempts or human labeling, making the collection of this dataset very efficient.  Furthermore, because these paired images are collected in the real world, they contain all of the real-world noise and artifacts that one would encounter in a realistic setting, avoiding the need to create such artifacts in simulation. 

\subsection{Supervision Transfer for Multi-modal Perception}
\label{sec:suptransfer}
\hl{In attempting the modality transfer described above, we observe the following}: different input modalities (\eg depth vs RGB) have complementary advantages. 
In other words, data that is difficult for computing successful grasps in one modality might not be as difficult for another modality, and vice versa.  For example, transparent and reflective objects are extremely difficult for depth-based grasping methods, due to the resulting noise or missing data in the depth image.  However, our experiments show that \hl{RGB-based grasping methods have a much higher success rate for these objects}.  On the other hand, highly textured objects may present difficulties for RGB grasping methods, \hl{but these textures do not manifest in depth-based methods.}

\begin{figure}[t]
    \centering
    \includegraphics[width=0.99\columnwidth]{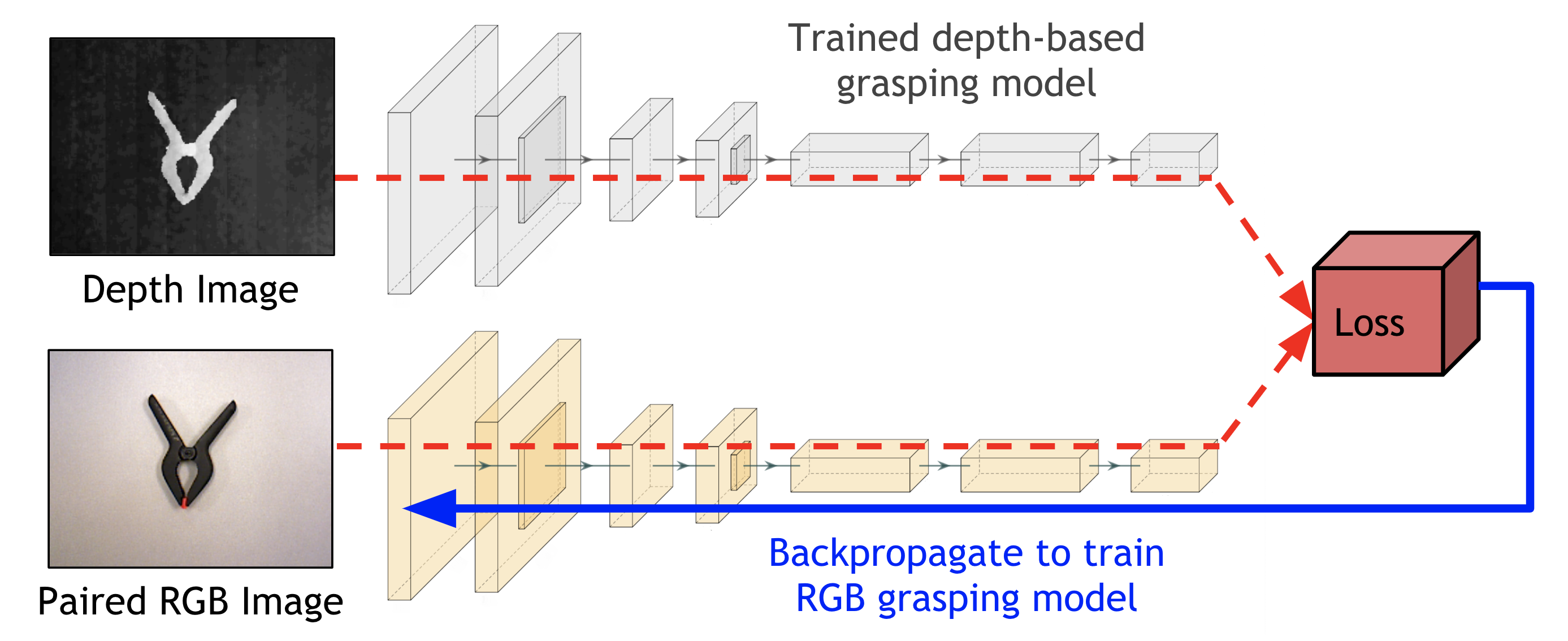}
    \caption{We train a grasp quality CNN that takes RGB input by supervising the loss of the network on the output of a trained depth model for paired, unlabeled RGB-D image data.}
    \label{fig:suptransfer}
\end{figure}

Based on this observation, we first filter our dataset $D$ into a new dataset $D'$ for which we expect the grasping method of modality $\mathcal{M}_d$ to perform well. 
In other words, for images $I_d \in D'$, the grasp score $G(q, I_d)$ should have a high correlation with the success of an executed grasp. In our case, because $I_d$ is a depth image, our filtered dataset $D'$ contains only images of opaque objects, for which depth-based grasping methods typically perform well.

We then train a grasping method for modality $\mathcal{M}_s$ (\eg RGB) using supervision transfer~\cite{gupta2016cross, hoffman2016cross, li2018cross} over dataset $D'$.
For each paired image $(I_d, I_s)$ in dataset $D'$, we compute the grasping score $G(q, I_d)$ for the modality $\mathcal{M}_d$\hl{. Because of our filtering, this grasp score is likely to be accurate}. We then train a method for computing the grasping score \hl{$G_\phi(q, I_s)$} of the second modality $\mathcal{M}_s$ \hl{using the grasp score from modality $\mathcal{M}_d$ as the grasp label; thus we define the loss to be}
\begin{align}
    \mathcal{L}(\phi) = ||G(q, I_d) - G_\phi(q, I_s)||^2
\end{align}
For paired images of dataset $D'$, we train the grasping method on the new modality $\mathcal{M}_s$ (\eg RGB) to output the same grasping score as the score output of the previous grasping method on the original modality $\mathcal{M}_d$ (\eg depth). This procedure is shown in Figure~\ref{fig:suptransfer}.

\hl{Because of the complementary nature of the two sensors, this grasping score function will often perform well on data that was originally filtered out of $D$ and not included in $D'$, even though $G_\phi(q, I_s$) was only trained on data from $D'$.
Specifically, we filter out transparent and reflective objects from $D'$ because depth-based grasping methods perform poorly on these objects.}
Nonetheless, the image-based grasping method $G_\phi(q, I_s)$ still performs well on  images of transparent and reflective objects, because the difference in appearance for these objects in the RGB modality is much smaller than the difference in appearance for these objects in the depth modality.  Our experiments confirm this to be the case.

Further, because the modalities are complementary, we show that we can get the best performance by combining the grasping scores from the two modalities. 
Although there are many potential ways to do this, we evaluate two possibilities.  The \hl{``early fusion'' approach for combining modalities is to transfer from a depth-based grasping network to a RGB-D grasping network (``RGBD-ST", see }\fig{fig:method-rgbdst}\hl{). RGBD-ST takes as input both depth and RGB modalities concatenated together.
For our second, ``late-fusion'' approach, we fuse the scores of each modality, averaging the outputs of the depth-based grasping network with a RGB-based grasping network trained using supervision transfer. We} define the multi-modal grasping score as
\begin{align}
    G_\phi(q, I_d, I_s) = \frac{1}{2} \cdot (G(q, I_d) + G_\phi(q, I_s))
\end{align}
This method is referred to below as ``RGBD-M" (\hl{see} \fig{fig:method-rgbdm}). Both of these approaches share the benefits that they represent multi-modal grasping methods that were trained from a depth-based grasping method only using paired RGB and depth images, without requiring real grasp attempts or human labels.

\subsection{Implementation of Supervision Transfer}
\label{sec:suptransfer_impl}

\begin{figure*}[ht]
    \centering
    \begin{subfigure}[t]{0.18\textwidth}
        \centering
        \includegraphics[scale=0.75]{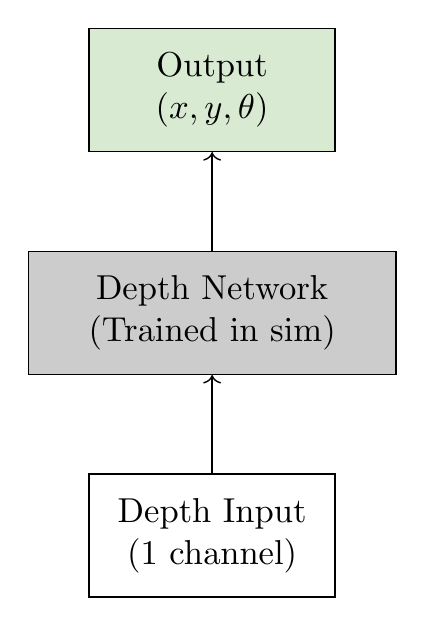}
        \caption{FC-GQCNN \cite{satish2019policy}}
        \label{fig:method-fcgqcnn}
    \end{subfigure}%
    ~
    \begin{subfigure}[t]{0.18\textwidth}
        \centering
        \includegraphics[scale=0.75]{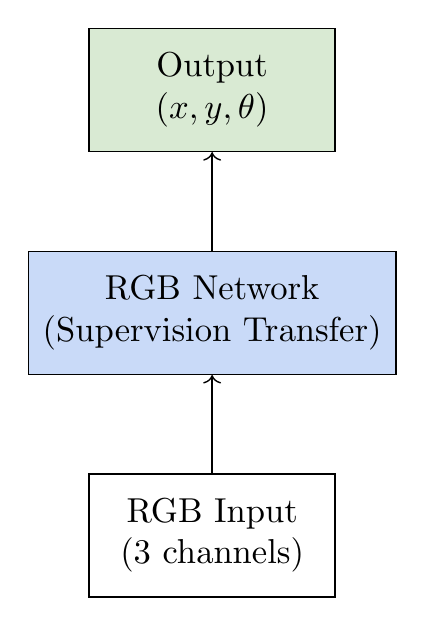}
        \caption{RGB-ST}
        \label{fig:method-rgbst}
    \end{subfigure}%
    ~ 
    \begin{subfigure}[t]{0.18\textwidth}
        \centering
        \includegraphics[scale=0.75]{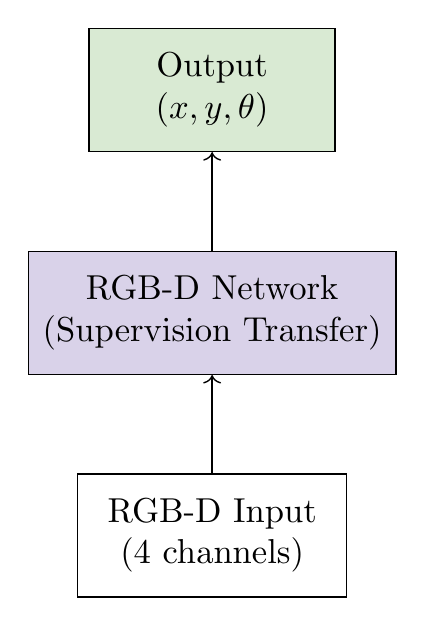}
        \caption{RGBD-ST}
        \label{fig:method-rgbdst}
    \end{subfigure}
    ~
    \begin{subfigure}[t]{0.34\textwidth}
        \centering
        \includegraphics[scale=0.75, trim={0 0 0 6pt}, clip]{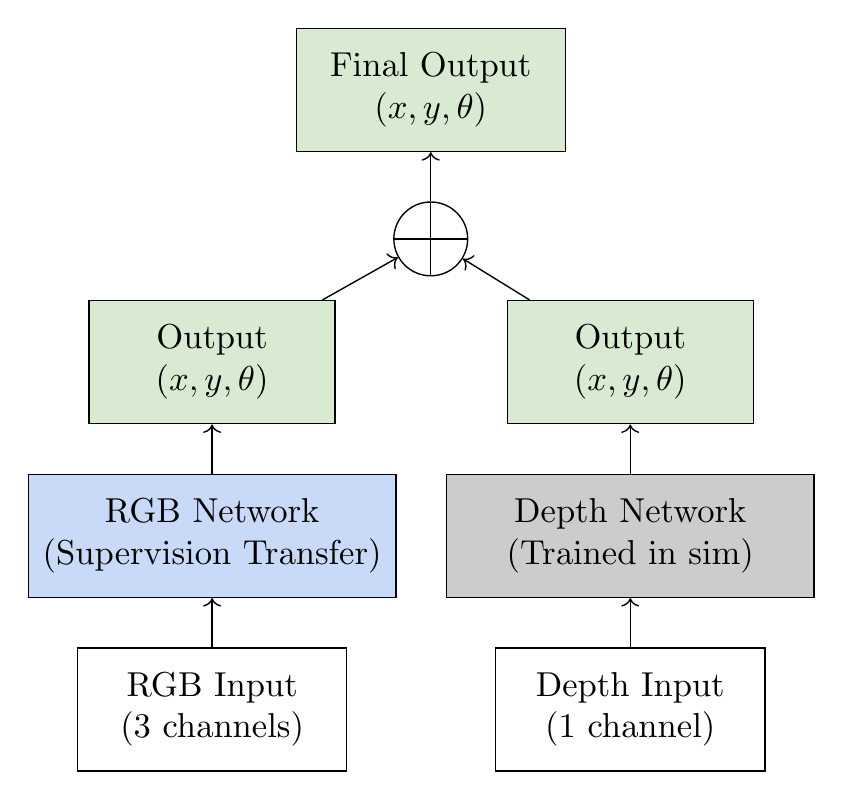}
        \caption{RGBD-M}
        \label{fig:method-rgbdm}
    \end{subfigure}
    \caption{\hl{Diagrams of the four methods evaluated in this work. We compare against FC-GQCNN }\cite{satish2019policy}\hl{, which takes a depth image as input and outputs dense grasp scores over image coordinates $x,y$ and rotation $\theta$ about the depth axis. RGB-ST and RGBD-ST are both trained using supervision transfer, but differ in the input they accept (3-channel RGB or 4-channel RGB-D input). RGBD-M takes the outputs of the RGB and Depth networks and averages them to produce the final output.}}
    \label{fig:methods}
\end{figure*}

Our supervision transfer formulation is agnostic to the specific grasping method or representation we use for grasping in modality $\mathcal{M}_d$.
For this work, we use the Fully Convolutional Grasp Quality CNNs (FC-GQCNN) representation as the pre-trained depth model from Satish \etal\cite{satish2019policy}, although other depth-based grasping methods could equivalently be used.

FC-GQCNN learns a function $G(q_d, I_d)$ which predicts a grasp success rate for each grasp $q_d$ based on a depth image $I_d$.  In FC-GQCNN, grasps $q_d$ are parameterized as $q_d = (x, y, \theta, z)$, where $x$ and $y$ are horizontal planar coordinates designating the desired grasp point of the gripper, $z$ is the grasp depth relative to the camera, and $\theta$ is the clockwise rotation angle of the gripper about the vertical $z$ axis.  \hl{FC-GQCNN takes as input just a single depth image $I_d$ and outputs a 4-dimensional tensor of grasping scores, producing one score per binned ($x$,$y$,$z$) position as well as binned orientation coordinates $\theta$}. FC-GQCNN is designed to be fully convolutional in order to output dense predictions $G(q_d, I_d)$ across the entire depth image. \hl{Our methods, shown in Figure}~\ref{fig:methods}, \hl{use a similarly dense ($x$, $y$) output and the same output angular encoding $\theta$.}

We wish to use the output of FC-GQCNN to train an image-based grasping method  $G(q, I_s)$.  Because the image modality does not have access to depth information, for image-based grasping we change the grasping parameterization to just $q = (x, y, \theta)$, without including a parameter for the grasp depth $z$.  With this specification, each grasp starts at an approach height and moves down until it makes contact with either the table or an object before closing the gripper.  Due to the difference in grasp representations, we modify our loss slightly, to be:
\begin{align}
    \mathcal{L}(\phi; q, I_d, I_s) = ||\max_z G((q, z), I_d) - G_\phi(q, I_s)||^2
\end{align}
where $(q, z)$ is the concatenation of $z$ to a grasp $q = (x, y, \theta)$ to form the new grasp representation $(x, y, \theta, z)$. In other words, to compute the target grasp score for some grasp $q = (x, y, \theta)$, we append various depths $z$ to form a depth-based grasp parameterization $(x, y, z, \theta)$; for each of these grasp parameterizations we can compute the depth-based grasping score $G((q, z), I_d)$ using our depth-based grasping method (e.g. FC-GQCNN).  We then compute the maximum grasp score over the values of $z$ to obtain $\max_z G((q, z), I_d)$.

The network architecture that we use for image-based grasping is very similar to the architecture used in FC-GQCNN for depth-based grasping (\hl{see Appendix A}).  The only modification that we make is that we modify the first layer to accept a 3-channel RGB input rather than a 1-channel depth input. This is accomplished by adding an extra dimension to the first layer convolutional filters. In some of the experiments, we will alternatively use an RGB-D grasping network (``RGBD-ST"), in which case we modify the first layer to accept a 4-channel input, in a similar manner.

\section{Experimental Setup}
\label{sec:exp_setup}

Following the reproducibility guidelines for grasping research as presented in~\cite{mahler2018guest}, we describe our experimental setup and protocols below.

\subsection{Physical Components}

We use an ASUS Xtion Pro Live RGB-D sensor, fixed 0.7 m above and pointing down towards the workspace (see \fig{fig:dataset}).
Robot experiments were performed on a 7 DOF Rethink Robotics Sawyer robot equipped with an electric parallel jaw gripper, though our method can be applied to other robots and end-effectors.
The robot's workspace is an approximately 0.65 m $\times$ 0.38 m area that is reachable by the robot with a vertical grasp. 
Aluminum extrusions enclose the workspace to prevent objects from rolling or sliding out of the space.  

All experiments and network training were performed on an Ubuntu 16.04 machine with an NVIDIA GTX 1080 Ti GPU, a 2.1 GHz Intel Xeon CPU, and 32 GB RAM allocated per job. Grasp planning was implemented using off-the-shelf MoveIt! software.

\subsection{Training the Network}

We first collected a set of 100 opaque objects from home and office retail stores.
Using the ASUS Xtion Pro Live RGB-D sensor fixed above the workspace, we captured 200 paired RGB-D training images and 50 paired validation images of the objects in varying amounts of clutter and with lighting conditions ranging from standard office illuminance (approx. 500 lux) to dimmed illuminance (approx. 175 lux).
We resized the images to account for differences between our sensor's intrinsic parameters and those of the pretrained FC-GQCNN model.
To increase the amount of training data and improve domain robustness, we applied spatial augmentations (\eg random rotations and flips) and color-based augmentations (\eg hue, brightness, and contrast), generating approximately 20k paired training images.
This image dataset is available at the URL in the abstract.

The network architecture was implemented in Python using Tensorflow and Keras. 
The RGB or RGB-D network's weights were randomly initialized, and the model was trained to convergence using an Adam optimizer with cross-entropy loss~\cite{kahn2018self, kalashnikov2018qt}. 
We experimented with mean squared error loss, but it performed worse in initial experiments. 
The loss was supervised from the output of FC-GQCNN, taking the maximum over all values of $z$ as discussed in \sect{sec:suptransfer_impl}. Hyperparameters are provided in Appendix C.

\subsection{Test Objects}

\begin{figure*}[t]
    \vspace{2pt}
    \centering
    \begin{subfigure}[b]{0.1875\textwidth}
        \centering
        \includegraphics[width=0.99\textwidth]{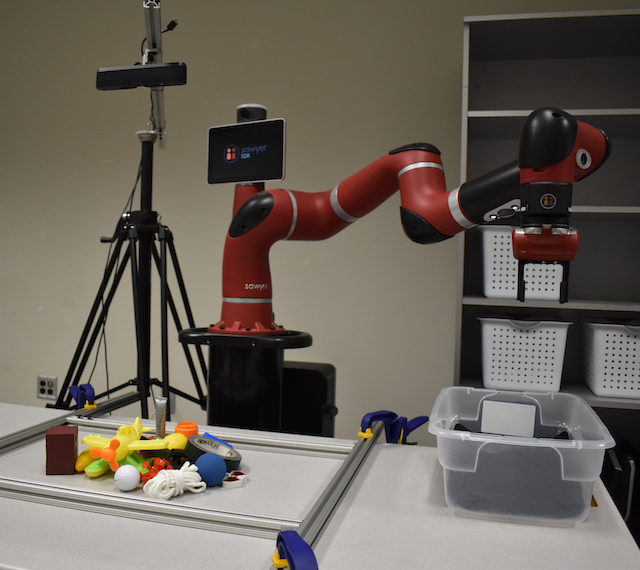}
    \end{subfigure}
    \begin{subfigure}[b]{0.25\textwidth}
        \includegraphics[width=0.99\textwidth]{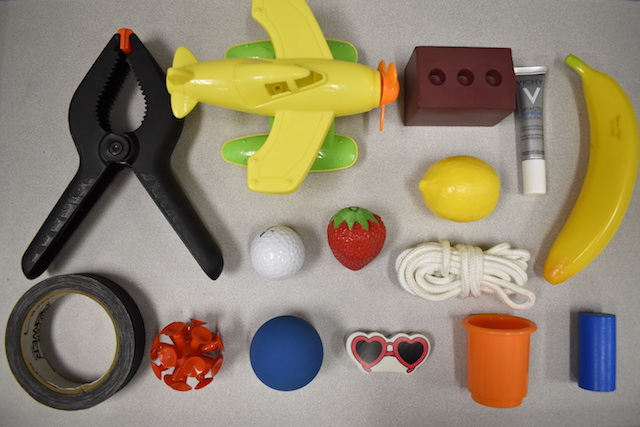}
    \end{subfigure}
    \begin{subfigure}[b]{0.25\textwidth}
        \includegraphics[width=0.99\textwidth]{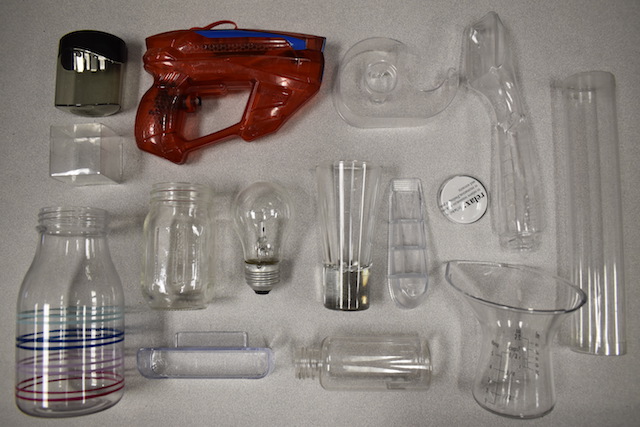}
    \end{subfigure}
    \begin{subfigure}[b]{0.25\textwidth}
        \includegraphics[width=0.99\textwidth]{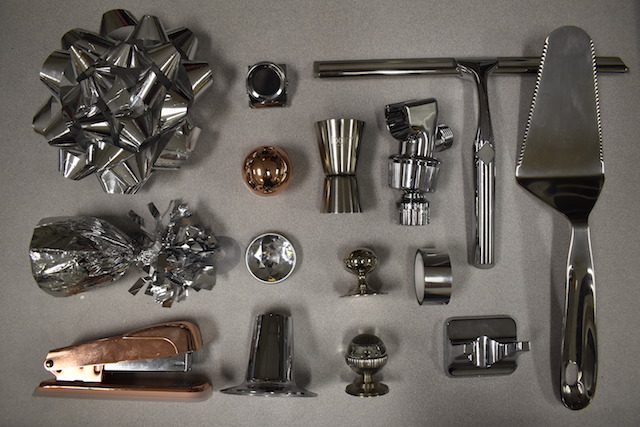}
    \end{subfigure}
\caption{Data collection setup and example images from the dataset of all three object types.}
\label{fig:dataset}
\end{figure*}

We collected objects distinct from the training objects to form three sets of 15 test objects each, one set per category (see \fig{fig:dataset}).
For the opaque object set, we primarily use YCB~\cite{calli2015benchmarking} objects that fit within the 5 cm stroke width of our gripper. 
We collected our own transparent and specular object sets due to the lack of existing benchmark sets for these categories. 

Following typical procedures for grasping evaluations~\cite{mahler2017dex, viereck2017learning}, we remove bias related to object pose through the following procedure: objects are shaken in a box and then emptied onto the robot's workspace for each grasp attempt. This procedure is used for both isolated object grasping as well as for grasping in clutter.

\section{Experimental Results}
\label{sec:results}

We design experiments to answer the following questions:
\begin{itemize}
    \item To what extent can supervision transfer be used to grasp objects from new modalities (e.g. depth to RGB)?
    \item To what extent can supervision transfer from depth to RGB be used to learn to grasp transparent and reflective objects?
    \item Do the depth and image modalities complement each other? That is, will combining both modalities outperform either modality alone?
\end{itemize}

Note that grasping performance is not directly comparable with previous work like FC-GQCNN~\cite{satish2019policy} as we use a different robot, gripper, and depth sensor.

\begin{figure*}[h]
    \centering
    \begin{subfigure}[b]{0.49\textwidth}
        \centering
        \includegraphics[height=3cm, keepaspectratio]{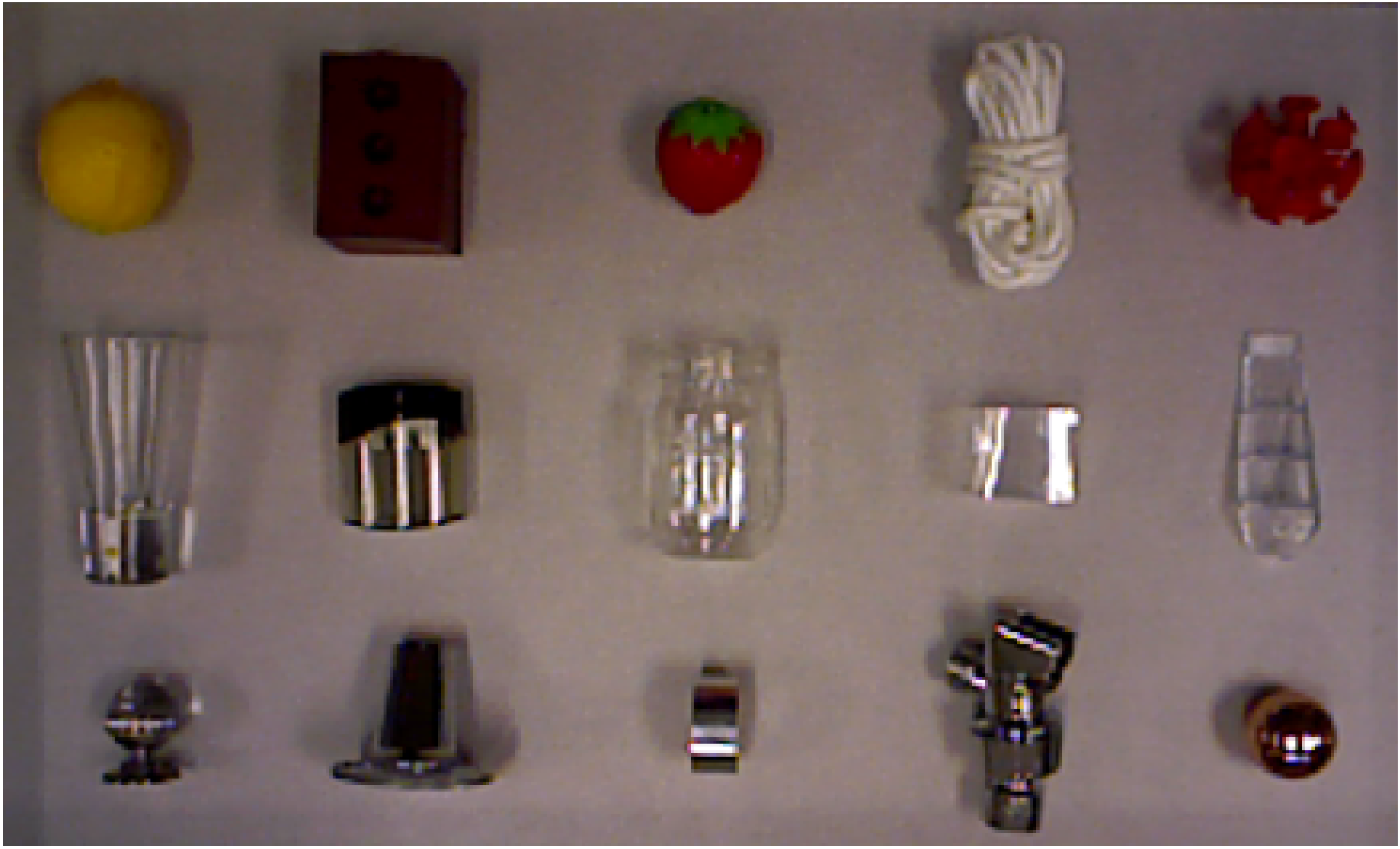}
        \caption*{RGB image}
    \end{subfigure}
    \vspace*{3px}
    \begin{subfigure}[b]{0.49\textwidth}
        \centering
        \includegraphics[height=3cm]{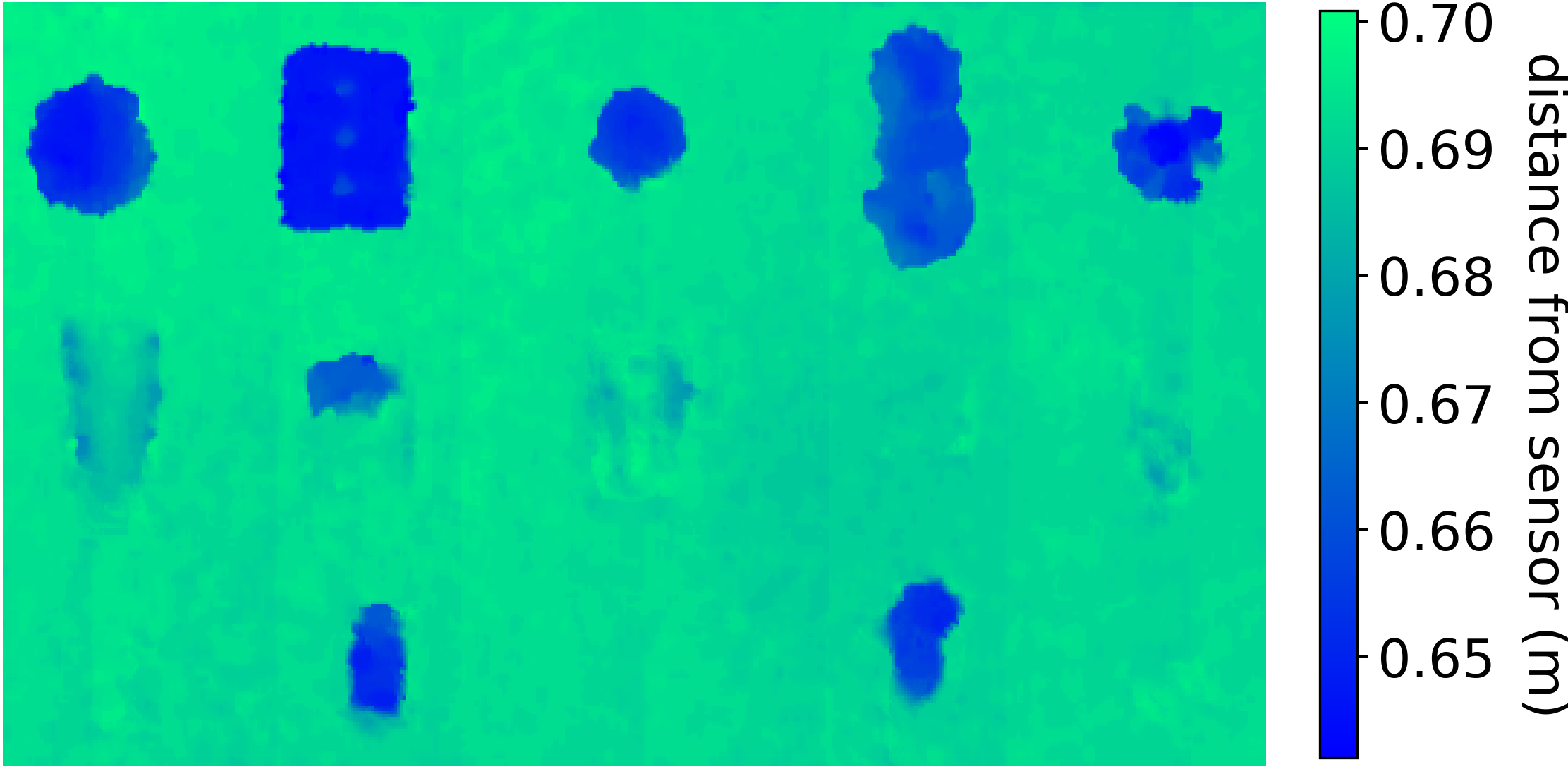}
        \caption*{Depth image}
    \end{subfigure}
    \vspace*{1px}
    \begin{subfigure}[b]{0.245\textwidth}
        \centering
        \includegraphics[width=0.99\textwidth]{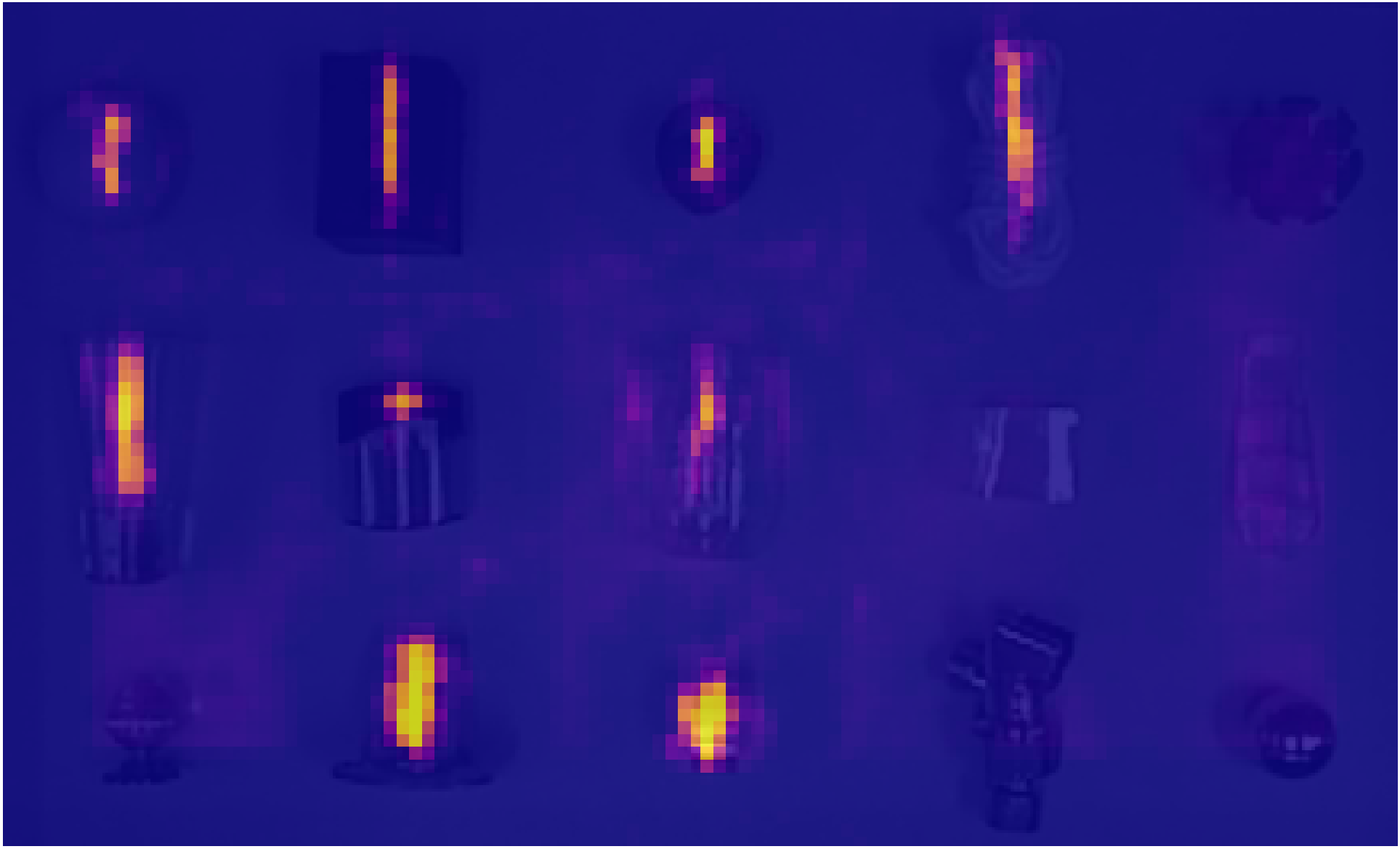}
        \caption*{FC-GQCNN}
    \end{subfigure}
    \begin{subfigure}[b]{0.245\textwidth}
        \centering
        \includegraphics[width=0.99\textwidth]{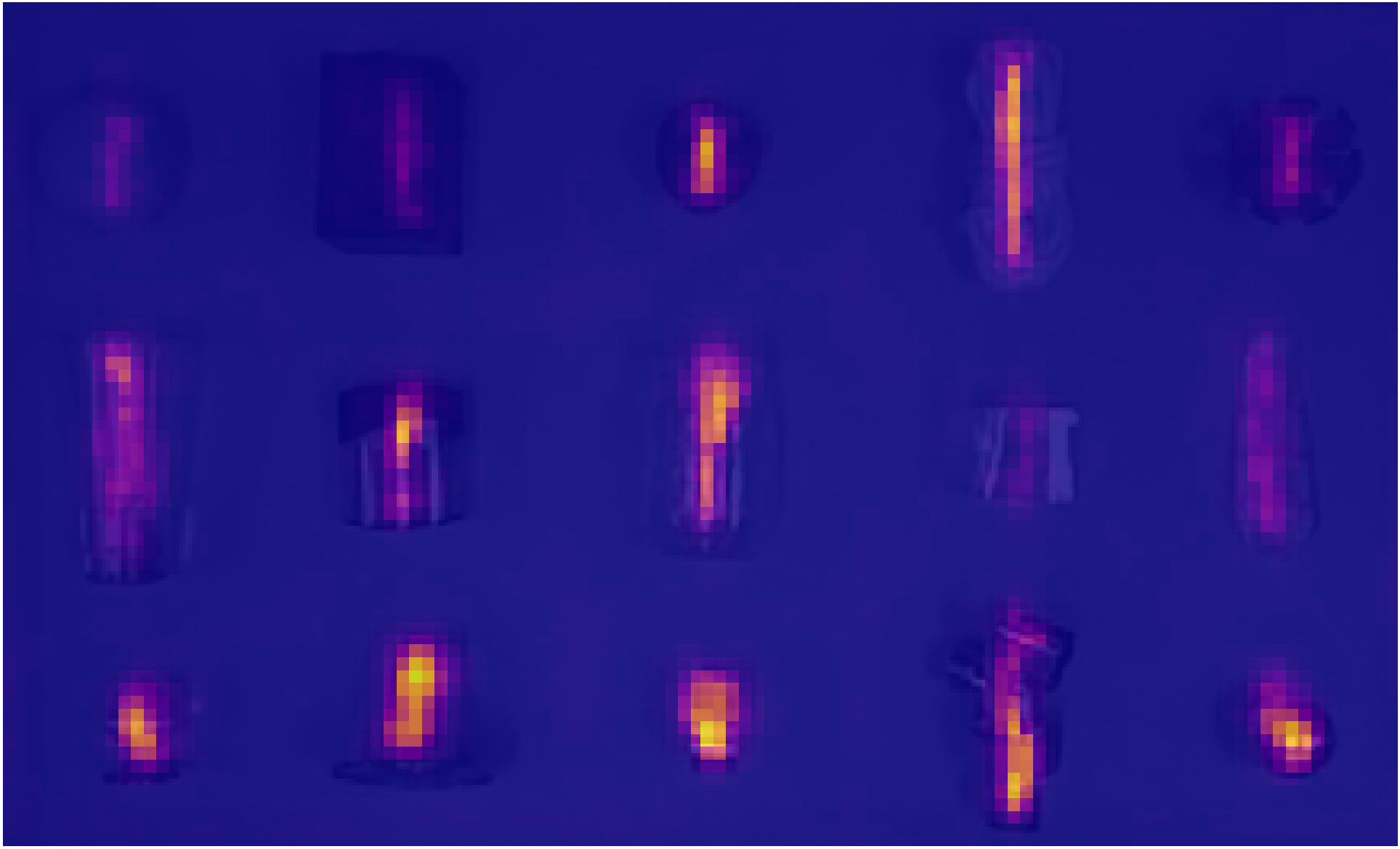}
        \caption*{RGB-ST}
    \end{subfigure}
    \begin{subfigure}[b]{0.245\textwidth}
        \centering
        \includegraphics[width=0.99\textwidth]{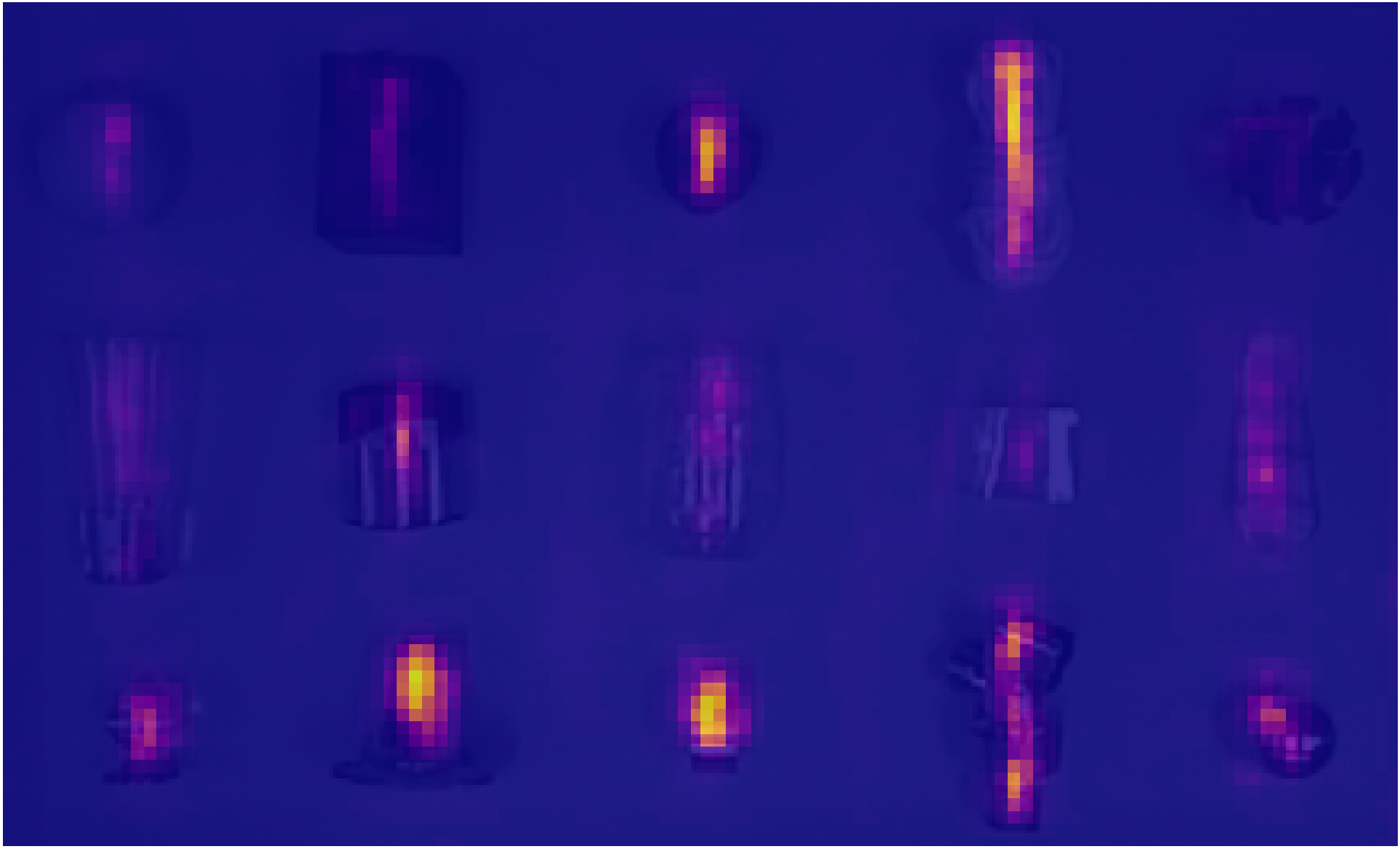}
        \caption*{RGBD-ST}
    \end{subfigure}
    \begin{subfigure}[b]{0.245\textwidth}
        \centering
        \includegraphics[width=0.99\textwidth]{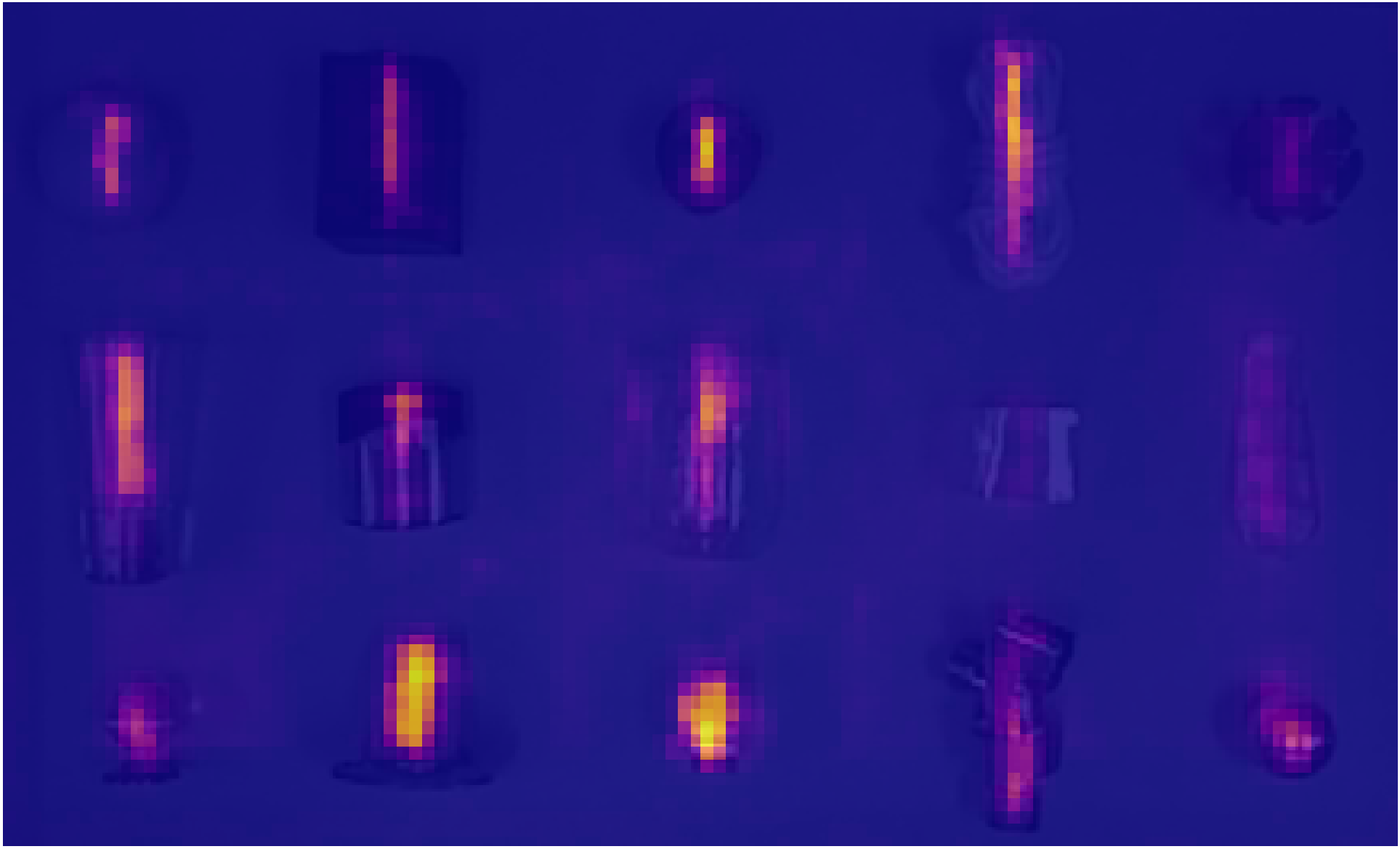}
        \caption*{RGBD-M}
    \end{subfigure}
    \begin{subfigure}[b]{0.99\textwidth}
        \centering
        \includegraphics[width=0.60\textwidth]{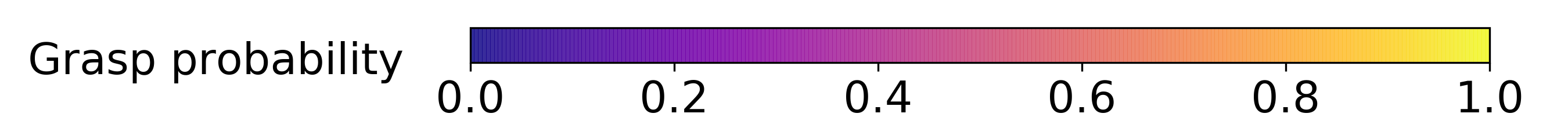}
    \end{subfigure}
    \caption{Probability heatmaps of grasping across methods for the max grasp score of a grasp with fingertips horizontal to the image, centered at the each pixel. Objects from each set are arrayed horizontally such that the top row is opaque objects, the next transparent, and the final one specular. }
    \label{fig:heatmap}
    \vspace{-2pt}
\end{figure*}

\subsection{Multi-modal Perception}
\label{sec:multimodal_perception}
We evaluate whether multi-modal perception that combines depth and RGB data is better than uni-modal perception using either depth or RGB data alone.  We refer to our method for Depth-to-RGB supervision transfer, described in Sections~\ref{sec:suptransfer} and~\ref{sec:suptransfer_impl}, as ``RGB-ST" (see \fig{fig:method-rgbst}).  

We evaluate two approaches to multi-modal perception, \hl{both of which are described in} Sections~\ref{sec:suptransfer} and~\ref{sec:suptransfer_impl}. 
The first \hl{``early-fusion"} approach uses supervision transfer to directly train an RGB-D grasp prediction network from a depth-based network, called ``RGBD-ST" (\hl{see }\fig{fig:method-rgbdst}).
The second \hl{``late-fusion"} approach involves taking the mean of the outputs of an RGB-only network and a depth-based network. 
Specifically, we take the mean of the RGB-ST and FC-GQCNN grasping networks; we call this multi-modal method ``RGBD-M" (\hl{see }\fig{fig:method-rgbdm}). 

The results are shown in Table~\ref{table:indiv_eval}.  
RGBD-ST and RGBD-M both significantly outperform depth-only grasping (FC-GQCNN) on transparent and specular objects, while maintaining comparable performance on opaque objects. 

We also see that the multi-modal methods perform similarly to the RGB-based grasping method (RGB-ST) on opaque and transparent objects, but outperform this method on specular objects. These results support the notion that combining both RGB and depth modalities gives better grasping performance than using either modality alone.

\begin{table}[h]
    \centering
    \caption{Isolated object grasping, averaged over five trials}
    \label{table:indiv_eval}
    \begin{tabular}{l ccc ccc}
      \toprule
        Method & Opaque & Transparent & Specular \\
        \midrule
        FC-GQCNN$^*$ & 
            $\mathbf{0.92\pm0.06}$ & $0.40\pm0.08$ & $0.48\pm0.17$\\
        RGB-ST$^\dagger$ & 
            $0.89\pm0.04$ & $0.79\pm0.09$ & $0.71\pm0.04$\\
        RGBD-ST$^\dagger$ & 
            $0.91\pm0.06$ & $0.77\pm0.08$ & $\mathbf{0.83\pm0.04}$ \\
        RGBD-M$^\dagger$ & 
            $0.91\pm0.14$ & $\mathbf{0.85\pm0.06}$ & $0.81\pm0.07$\\
      \bottomrule
    \end{tabular}
    \vspace{7pt}
    \caption*{\footnotesize $^*$Trained on simulated grasps \\ $^\dagger$Trained on simulated grasps and opaque object images \\}
    \vspace{-3em}
\end{table}

\subsection{Grasping in Clutter}

We also evaluated our methods for grasping in clutter, as this is important for robots in various cluttered environments like homes and warehouses.   
The same test objects used in isolated object grasping were used for clutter experiments. 
Five trials of grasping in clutter were conducted for each object category. 
Following the procedure from Viereck \etal\cite{viereck2017learning}, a trial concluded after all objects were successfully grasped, 3 consecutive failed grasp attempts occurred, or all objects were outside the workspace.

To prevent a network from getting repeatedly stuck on attempting a bad but highly rated grasp, we randomly sample a 0.2m square crop of the input image and select the grasp location within that region with the maximum predicted success probability.
All methods including baselines performed similarly or worse without this sampling (see Appendix B).
Crops whose grasp probabilities all fall below below a threshold are discarded and resampled to avoid attempting grasps based on noisy sensor readings. 

\begin{table}[h]
    \centering
    \caption{Grasping in clutter, averaged over five trials}
    \label{table:clutter_eval}
    \begin{tabular}{l cc cc cc}
      \toprule
        Method 
            & Opaque
            & Transparent
            & Specular \\
        \midrule
        FC-GQCNN$^*$
            & $0.84\pm0.06$
            & $0.23\pm0.21$
            & $0.35\pm0.16$ \\
        RGB-ST$^\dagger$
            & $0.77\pm0.11$
            & $\mathbf{0.67\pm0.10}$
            & $\mathbf{0.68\pm0.12}$ \\
        RGBD-ST$^\dagger$
            & $0.86\pm0.09$
            & $0.67\pm0.27$
            & $0.35\pm0.10$ \\
        RGBD-M$^\dagger$ 
            & $\mathbf{0.97\pm0.15}$
            & $0.51\pm0.32$
            & $0.63\pm0.12$ \\
      \bottomrule
    \end{tabular}
    \vspace{3pt}
    \caption*{\footnotesize $^*$Trained on simulated\ grasps \\ $^\dagger$Trained on simulated\ grasps and opaque object images \\}
    \vspace{-2em}
\end{table}

The results are shown in Table~\ref{table:clutter_eval}.
The results from grasping in clutter corroborate the result of isolated grasping. 
All methods perform well on opaque objects, although RGBD-M (averaging the output of depth-only grasping and RGB-only grasping networks) performs slightly better than the others.  On non-opaque objects (e.g. transparent and specular), FC-GQCNN (e.g. depth-only grasping) performs poorly.

Table~\ref{table:clutter_eval} shows that RGB-ST (RGB-only grasping) and RGBD-M (averaging the output of depth-only grasping and RGB-only grasping networks) perform well across all three object categories.  We note that, despite averaging across five trials, the results of grasping in clutter have relatively high variance and should be considered accordingly.  Overall, our main conclusions are similar to that of isolated object grasping from Section~\ref{sec:multimodal_perception}: depth-only grasping performs poorly on transparent and specular objects; with supervision transfer, we can obtain a method that performs much better on grasping transparent and specular objects while maintaining similar performance on opaque objects.
This method requires only paired RGB and depth images for training and does not require any real grasp attempts or human annotations, \hl{other than the simulated depth rendering data that was used to train the original FC-GQCNN}~\cite{satish2019policy} \hl{depth-based grasping method}.

\subsection{Lighting Variation Experiments}

We note that domain shifts like lighting can be a problem for RGB methods, as mentioned in previous work~\cite{mahler2018guest}. To enable our method to be robust to lighting variations, our training images were collected with slight lighting variations, and we applied color-based augmentations like brightness and contrast.

\begin{figure}[h]
    \centering
    \begin{subfigure}[b]{0.49\columnwidth}
        \centering
        \includegraphics[width=0.99\linewidth]{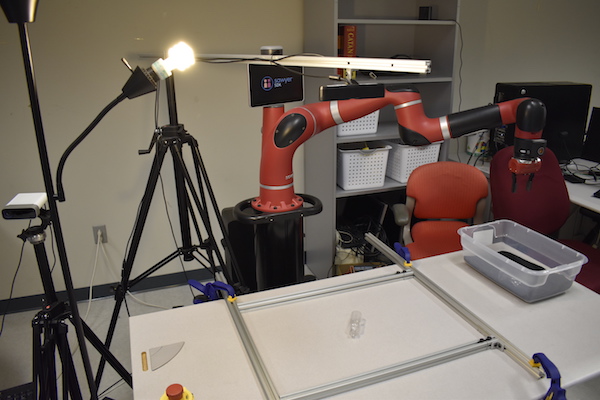}
        \caption{Lighting setup.}
        \label{fig:lighting1}
    \end{subfigure}
    \begin{subfigure}[b]{0.49\columnwidth}
        \centering
        \includegraphics[width=0.99\linewidth]{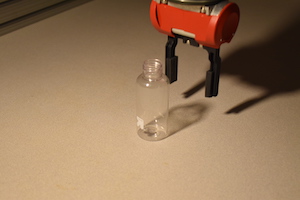}
        \caption{Extreme lighting.}
        \label{fig:lighting2}
    \end{subfigure}
    \caption{\hl{(a) Setup for lighting variation experiments. Lighting is controlled using the overhead lights and floor lamp. (b) Failure case in the extreme lighting condition. The method predicts the best grasp to be on the object's shadow.
    }}
\end{figure}

We conducted experiments to evaluate the robustness of the trained networks to lighting variations. We varied the lighting by moving a floor lamp around the robot workspace as shown in \fig{fig:lighting1} and performed the isolated object grasping experiments for RGBD-M. The additional lighting increased illumination to between 750 and 950 lux. With this variation in lighting, the RGBD-M network performed comparably, achieving grasp success rates of $0.81\pm0.12$ \hl{for transparent objects and $0.79\pm0.09$ for specular ones} (compare with \tbl{table:indiv_eval}).

However, we found that the network performed poorly under more drastic lighting changes, in which \hl{we turned off the overhead lights and reduced the height of  the floor light}, dropping illumination to approx.\ 175 lux and causing long object shadows to appear.  
\hl{In this case, grasp performance dropped to $0.52\pm0.18$ on transparent objects and $0.60\pm0.12$ for specular ones}. 
In such extreme lighting conditions, we observed the method predicting grasps on shadows for transparent objects (\hl{see }Fig.~\ref{fig:lighting2}). Such drastic lighting would not normally occur in structured applications like bin-picking.

\subsection{Failure Cases}

\begin{figure}[h]
    \centering
    \includegraphics[width=0.35\columnwidth]{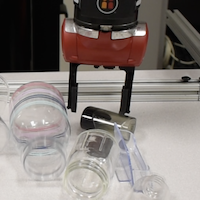}
    \vspace{0.3em}
    \includegraphics[width=0.35\columnwidth]{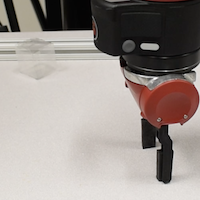}
    \includegraphics[width=0.35\columnwidth]{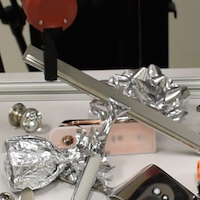}
    \includegraphics[width=0.35\columnwidth]{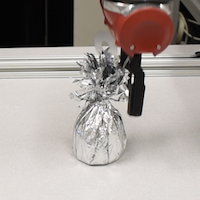}
    \caption{Examples of failure cases. (top left) Grasp does not account for transparent part of sharpener. (top right) Gripper fails to detect transparent plastic cube and grasps at table. (bottom left) Mass distribution of squeegee causes grasp to fail. (bottom right) Foil on top of balloon weight appears graspable but the gripper passes through.}
    \label{fig:fail}
\end{figure}

In this section we discuss the most frequent and notable failure cases from our experiments. 
This section covers failures due to our approach, as well as external factors.
Some examples of failure cases discussed in this section can be seen in Fig.~\ref{fig:fail} and the supplementary video. 

Methods that used the depth modality like RGBD-ST and RGBD-M at times selected grasps that were highly rated by the depth network, but did not sufficiently account for transparencies or specularities (Fig.~\ref{fig:fail}, top left).
Both the color-based and depth-based networks at times failed to distinguish very transparent objects from the workspace surface, though this was rare and occurred far less frequently than with FC-GQCNN (Fig.~\ref{fig:fail}, top right).  
Object mass distribution and deformability were not accounted for by our methods (Fig.~\ref{fig:fail}, bottom row).

A failure case external to the methods evaluated involved our gripper hardware.
Our parallel electric gripper has a relatively small stroke width, and is unable to execute pinch grasps with a 5cm opening width.
This limitation causes grasps on thin parts of objects to fail, because the fingertips do not completely come together.
While it is possible to adjust the fingertips to be closer together to enable pinch grasps, the opening width of the gripper would be reduced,  which would prevent the gripper from being able to grasp large objects.
This issue reduced performance across all methods and would likely be mitigated by other grippers.

Since our paper focused on static grasping, our method fails to grasp objects that start rolling due to perturbation in clutter. 
Others have investigated ways to address this issue using closed-loop control techniques like visual servoing~\cite{morrison2018closing}.

\section{Conclusion}
\label{sec:conclusion}

We present an approach for improving grasping on transparent and specular objects, for which existing depth-based grasping methods perform poorly. 
Our method transfers information learned by a depth-based grasping network to RGB or RGB-D networks, enabling multi-modal perception. 
Our method for supervision transfer requires only real-world paired depth and RGB images, and does not require any human labeling nor real-world grasp attempts. 
We explore two avenues to multi-modal perception and demonstrate that making use of the RGB modality outperforms depth-only grasping in isolated object grasping as well as grasping in clutter.
The method is extensible to other robots, environments, and end effectors. 
One potential direction for future work may be to adaptively weight predictions from different modalities instead of averaging them.
Another is applying transfer learning techniques to other, less similar modalities like haptics and tactile feedback. \hl{Combining different sensor modalities might also be useful in determining the appropriate grasp height for each object.}

While we are able to get improved performance without using any real grasping data, we believe that real grasps can be used to further improve the performance of the network. We are also interested in extending this work to other types of grasping, such as 6-DOF, multi-fingered, or suction grasping.









\bibliographystyle{IEEEtran.bst}
\bibliography{root}

\begin{thebibliography}{10}
\providecommand{\url}[1]{#1}
\csname url@rmstyle\endcsname
\providecommand{\newblock}{\relax}
\providecommand{\bibinfo}[2]{#2}
\providecommand\BIBentrySTDinterwordspacing{\spaceskip=0pt\relax}
\providecommand\BIBentryALTinterwordstretchfactor{4}
\providecommand\BIBentryALTinterwordspacing{\spaceskip=\fontdimen2\font plus
\BIBentryALTinterwordstretchfactor\fontdimen3\font minus
  \fontdimen4\font\relax}
\providecommand\BIBforeignlanguage[2]{{%
\expandafter\ifx\csname l@#1\endcsname\relax
\typeout{** WARNING: IEEEtran.bst: No hyphenation pattern has been}%
\typeout{** loaded for the language `#1'. Using the pattern for}%
\typeout{** the default language instead.}%
\else
\language=\csname l@#1\endcsname
\fi
#2}}

\bibitem{satish2019policy}
V.~Satish, J.~Mahler, and K.~Goldberg, ``On-policy dataset synthesis for
  learning robot grasping policies using fully convolutional deep networks,''
  \emph{IEEE Robotics and Automation Letters}, vol.~4, no.~2, pp. 1357--1364,
  2019.

\bibitem{morrison2018closing}
D.~Morrison, J.~Leitner, and P.~Corke, ``Closing the loop for robotic grasping:
  A real-time, generative grasp synthesis approach,'' in \emph{Proceedings of
  Robotics: Science and Systems}, Pittsburgh, Pennsylvania, June 2018.

\bibitem{gualtieri2016high}
M.~Gualtieri, A.~Ten~Pas, K.~Saenko, and R.~Platt, ``High precision grasp pose
  detection in dense clutter,'' in \emph{2016 IEEE/RSJ International Conference
  on Intelligent Robots and Systems (IROS)}.\hskip 1em plus 0.5em minus
  0.4em\relax IEEE, 2016, pp. 598--605.

\bibitem{ihrke2010transparent}
I.~Ihrke, K.~N. Kutulakos, H.~P. Lensch, M.~Magnor, and W.~Heidrich,
  ``Transparent and specular object reconstruction,'' in \emph{Computer
  Graphics Forum}, vol.~29, no.~8.\hskip 1em plus 0.5em minus 0.4em\relax Wiley
  Online Library, 2010, pp. 2400--2426.

\bibitem{hoffman2016cross}
J.~Hoffman, S.~Gupta, J.~Leong, S.~Guadarrama, and T.~Darrell, ``Cross-modal
  adaptation for rgb-d detection,'' in \emph{2016 IEEE International Conference
  on Robotics and Automation (ICRA)}.\hskip 1em plus 0.5em minus 0.4em\relax
  IEEE, 2016, pp. 5032--5039.

\bibitem{curless1995better}
B.~Curless and M.~Levoy, ``Better optical triangulation through spacetime
  analysis,'' in \emph{Proceedings of IEEE International Conference on Computer
  Vision}.\hskip 1em plus 0.5em minus 0.4em\relax IEEE, 1995, pp. 987--994.

\bibitem{6619203}
K.~{Maeno}, H.~{Nagahara}, A.~{Shimada}, and R.~{Taniguchi}, ``Light field
  distortion feature for transparent object recognition,'' in \emph{2013 IEEE
  Conference on Computer Vision and Pattern Recognition}, June 2013, pp.
  2786--2793.

\bibitem{wetzstein2011hand}
G.~Wetzstein, R.~Raskar, and W.~Heidrich, ``Hand-held schlieren photography
  with light field probes,'' in \emph{2011 IEEE International Conference on
  Computational Photography (ICCP)}.\hskip 1em plus 0.5em minus 0.4em\relax
  IEEE, 2011, pp. 1--8.

\bibitem{oberlintime}
J.~Oberlin and S.~Tellex, ``Time-lapse light field photography for perceiving
  transparent and reflective objects.''

\bibitem{alhwarin2014ir}
F.~Alhwarin, A.~Ferrein, and I.~Scholl, ``Ir stereo kinect: improving depth
  images by combining structured light with ir stereo,'' in \emph{Pacific Rim
  International Conference on Artificial Intelligence}.\hskip 1em plus 0.5em
  minus 0.4em\relax Springer, 2014, pp. 409--421.

\bibitem{chiu2011improving}
W.~W.-C. Chiu, U.~Blanke, and M.~Fritz, ``Improving the kinect by cross-modal
  stereo.''\hskip 1em plus 0.5em minus 0.4em\relax Citeseer.

\bibitem{mahler2018guest}
J.~Mahler, R.~Platt, A.~Rodriguez, M.~Ciocarlie, A.~Dollar, R.~Detry, M.~A.
  Roa, H.~Yanco, A.~Norton, J.~Falco, \emph{et~al.}, ``Guest editorial open
  discussion of robot grasping benchmarks, protocols, and metrics,'' \emph{IEEE
  Transactions on Automation Science and Engineering}, vol.~15, no.~4, pp.
  1440--1442, 2018.

\bibitem{lysenkov2013recognition}
I.~Lysenkov, V.~Eruhimov, and G.~Bradski, ``Recognition and pose estimation of
  rigid transparent objects with a kinect sensor,'' \emph{Robotics}, vol. 273,
  2013.

\bibitem{lysenkov2013pose}
I.~Lysenkov and V.~Rabaud, ``Pose estimation of rigid transparent objects in
  transparent clutter,'' in \emph{2013 IEEE International Conference on
  Robotics and Automation}.\hskip 1em plus 0.5em minus 0.4em\relax IEEE, 2013,
  pp. 162--169.

\bibitem{bohg2013data}
J.~Bohg, A.~Morales, T.~Asfour, and D.~Kragic, ``Data-driven grasp
  synthesis—a survey,'' \emph{IEEE Transactions on Robotics}, vol.~30, no.~2,
  pp. 289--309, 2013.

\bibitem{miller2003automatic}
A.~T. Miller, S.~Knoop, H.~I. Christensen, and P.~K. Allen, ``Automatic grasp
  planning using shape primitives,'' 2003.

\bibitem{ten2017grasp}
A.~ten Pas, M.~Gualtieri, K.~Saenko, and R.~Platt, ``Grasp pose detection in
  point clouds,'' \emph{The International Journal of Robotics Research},
  vol.~36, no. 13-14, pp. 1455--1473, 2017.

\bibitem{watkins2018multi}
D.~Watkins-Valls, J.~Varley, and P.~Allen, ``Multi-modal geometric learning for
  grasping and manipulation,'' \emph{arXiv preprint arXiv:1803.07671}, 2018.

\bibitem{saxena2008robotic}
A.~Saxena, J.~Driemeyer, and A.~Y. Ng, ``Robotic grasping of novel objects
  using vision,'' \emph{The International Journal of Robotics Research},
  vol.~27, no.~2, pp. 157--173, 2008.

\bibitem{jiang2011efficient}
Y.~Jiang, S.~Moseson, and A.~Saxena, ``Efficient grasping from rgbd images:
  Learning using a new rectangle representation,'' in \emph{2011 IEEE
  International Conference on Robotics and Automation}.\hskip 1em plus 0.5em
  minus 0.4em\relax IEEE, 2011, pp. 3304--3311.

\bibitem{doi:10.1177/0278364914549607}
\BIBentryALTinterwordspacing
I.~Lenz, H.~Lee, and A.~Saxena, ``Deep learning for detecting robotic grasps,''
  \emph{The International Journal of Robotics Research}, vol.~34, no. 4-5, pp.
  705--724, 2015. [Online]. Available:
  \url{https://doi.org/10.1177/0278364914549607}
\BIBentrySTDinterwordspacing

\bibitem{7139361}
J.~{Redmon} and A.~{Angelova}, ``Real-time grasp detection using convolutional
  neural networks,'' in \emph{2015 IEEE International Conference on Robotics
  and Automation (ICRA)}, May 2015, pp. 1316--1322.

\bibitem{pinto2016supersizing}
L.~Pinto and A.~Gupta, ``Supersizing self-supervision: Learning to grasp from
  50k tries and 700 robot hours,'' in \emph{2016 IEEE international conference
  on robotics and automation (ICRA)}.\hskip 1em plus 0.5em minus 0.4em\relax
  IEEE, 2016, pp. 3406--3413.

\bibitem{levine2018learning}
S.~Levine, P.~Pastor, A.~Krizhevsky, J.~Ibarz, and D.~Quillen, ``Learning
  hand-eye coordination for robotic grasping with deep learning and large-scale
  data collection,'' \emph{The International Journal of Robotics Research},
  vol.~37, no. 4-5, pp. 421--436, 2018.

\bibitem{depierre2018jacquard}
A.~Depierre, E.~Dellandr{\'e}a, and L.~Chen, ``Jacquard: A large scale dataset
  for robotic grasp detection,'' in \emph{2018 IEEE/RSJ International
  Conference on Intelligent Robots and Systems (IROS)}.\hskip 1em plus 0.5em
  minus 0.4em\relax IEEE, 2018, pp. 3511--3516.

\bibitem{doi:10.1177/0278364919859066}
\BIBentryALTinterwordspacing
D.~Morrison, P.~Corke, and J.~Leitner, ``Learning robust, real-time, reactive
  robotic grasping,'' \emph{The International Journal of Robotics Research},
  vol.~0, no.~0, p. 0278364919859066, 0. [Online]. Available:
  \url{https://doi.org/10.1177/0278364919859066}
\BIBentrySTDinterwordspacing

\bibitem{mahler2017dex}
J.~Mahler, J.~Liang, S.~Niyaz, M.~Laskey, R.~Doan, X.~Liu, J.~A. Ojea, and
  K.~Goldberg, ``Dex-net 2.0: Deep learning to plan robust grasps with
  synthetic point clouds and analytic grasp metrics,'' \emph{arXiv preprint
  arXiv:1703.09312}, 2017.

\bibitem{gupta2016cross}
S.~Gupta, J.~Hoffman, and J.~Malik, ``Cross modal distillation for supervision
  transfer,'' in \emph{Proceedings of the IEEE conference on computer vision
  and pattern recognition}, 2016, pp. 2827--2836.

\bibitem{li2018cross}
G.~Li, Y.~Gan, H.~Wu, N.~Xiao, and L.~Lin, ``Cross-modal attentional context
  learning for rgb-d object detection,'' \emph{IEEE Transactions on Image
  Processing}, vol.~28, no.~4, pp. 1591--1601, 2018.

\bibitem{deng2009imagenet}
J.~Deng, W.~Dong, R.~Socher, L.-J. Li, K.~Li, and L.~Fei-Fei, ``Imagenet: A
  large-scale hierarchical image database,'' in \emph{2009 IEEE conference on
  computer vision and pattern recognition}.\hskip 1em plus 0.5em minus
  0.4em\relax Ieee, 2009, pp. 248--255.

\bibitem{kahn2018self}
G.~Kahn, A.~Villaflor, B.~Ding, P.~Abbeel, and S.~Levine, ``Self-supervised
  deep reinforcement learning with generalized computation graphs for robot
  navigation,'' in \emph{2018 IEEE International Conference on Robotics and
  Automation (ICRA)}.\hskip 1em plus 0.5em minus 0.4em\relax IEEE, 2018, pp.
  1--8.

\bibitem{kalashnikov2018qt}
D.~Kalashnikov, A.~Irpan, P.~Pastor, J.~Ibarz, A.~Herzog, E.~Jang, D.~Quillen,
  E.~Holly, M.~Kalakrishnan, V.~Vanhoucke, \emph{et~al.}, ``Qt-opt: Scalable
  deep reinforcement learning for vision-based robotic manipulation,''
  \emph{arXiv preprint arXiv:1806.10293}, 2018.

\bibitem{calli2015benchmarking}
B.~Calli, A.~Walsman, A.~Singh, S.~Srinivasa, P.~Abbeel, and A.~M. Dollar,
  ``Benchmarking in manipulation research: The ycb object and model set and
  benchmarking protocols,'' \emph{arXiv preprint arXiv:1502.03143}, 2015.

\bibitem{viereck2017learning}
U.~Viereck, A.~t. Pas, K.~Saenko, and R.~Platt, ``Learning a visuomotor
  controller for real world robotic grasping using simulated depth images,''
  \emph{arXiv preprint arXiv:1706.04652}, 2017.

\end{thebibliography}




\begin{figure*}[t]
    \centering
    \includegraphics[width=0.99\textwidth]{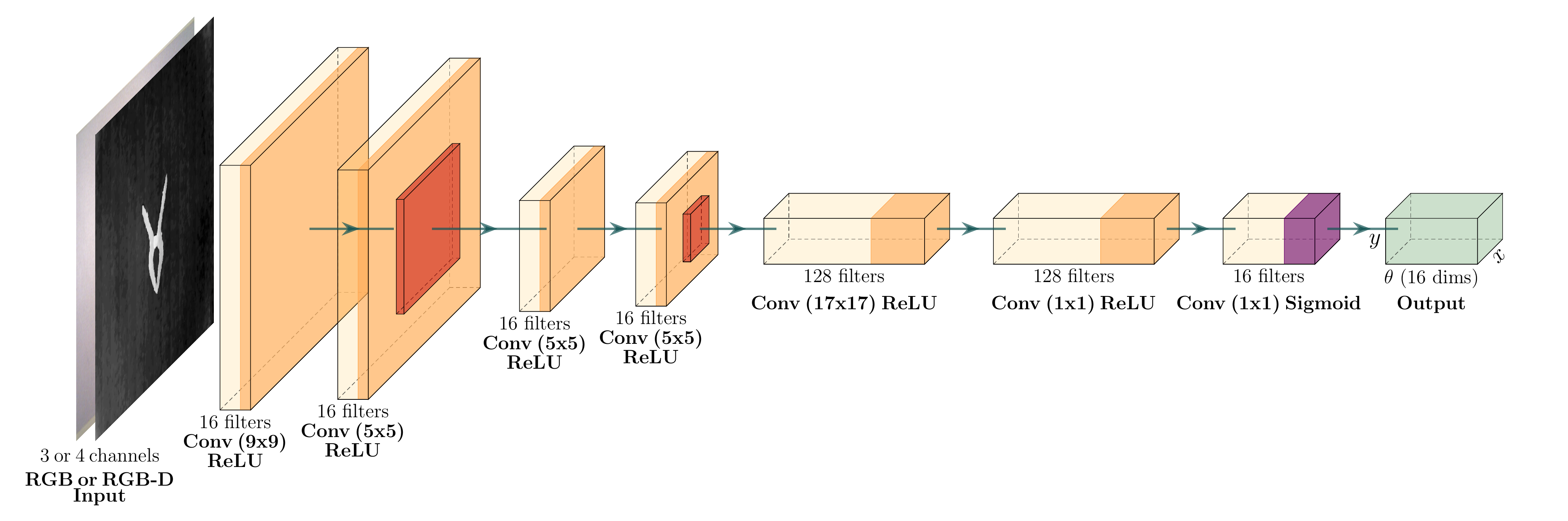}
    \caption{Architecture diagram for supervision transfer networks, adapted from the FC-GQCNN \cite{satish2019policy} architecture. The input can be either 3-channel RGB input or 4-channel RGB-D input. The output is a 3D array of grasp quality scores over image coordinates $x, y$ and rotation $\theta$ about the depth axis, discretized into 16 bins. The orange color accents correspond to ReLU activations and purple corresponds to sigmoid activation. The red layers are max pooling layers.}
    \label{fig:architecture}
\end{figure*}

\newpage



\begin{appendices}



\section{Network Architecture}
\label{appendix:architecture}

Fig.~\ref{fig:architecture} illustrates the architecture of the networks trained with supervision transfer.

\section{Evaluations without random cropping}
\label{appendix:nocrop}

Table \ref{table:clutter_eval_nocrop} provides results for grasping in clutter without random cropping.  

\begin{table}[h]
    \centering
    \caption{Performance on grasping in clutter by method without random cropping, averaged over five trials}
    \label{table:clutter_eval_nocrop}
    \begin{tabular}{l cc cc cc}
      \toprule
        Method 
            & Opaque
            & Transparent
            & Specular \\
        \midrule
        \vspace{0.6em}
        FC-GQCNN$^*$
            & $0.95\pm0.05$
            & $0.26\pm0.25$
            & $0.35\pm0.23$ \\
        RGB-ST$^\dagger$
            & $0.77\pm0.10$
            & $0.77\pm0.15$
            & $0.68\pm0.15$ \\
        RGBD-ST$^\dagger$
            & $0.62\pm0.26$
            & $0.67\pm0.19$
            & $0.75\pm0.08$ \\
        RGBD-M$^\dagger$ 
            & $0.75\pm0.13$
            & $0.60\pm0.18$
            & $0.47\pm0.28$ \\
      \bottomrule
    \end{tabular}
    \vspace{\baselineskip}
    \caption*{\footnotesize $^*$Trained on simulated\ grasps \\ $^\dagger$Trained on simulated\ grasps and opaque object images \\}
    \vspace{-1em}
\end{table}


\noindent Random cropping refers to sampling a 0.2m square crop from the input image and choosing the grasp with the highest probability from within the crop. Crops which have do not have any objects in them, as determined by whether the max grasp probability within the crop falls below a hand-defined threshold, are discarded and a new crop is sampled. This procedure helps prevent networks from repeatedly choosing highly rated false positive grasps. However, the cropping threshold must be tuned based on the performance of the grasping network.  For our experiments, we used a threshold of 0.4.

\pagebreak
\section{Hyperparameters}
\label{appendix:hyperparams}

\noindent Hyperparameters for networks trained with supervision transfer are:

\begin{itemize}
    \item Learning rate: 1e-05
    \item Batch size: 64
    \item Number of rotation augmentations per image: 32
    \item Loss: Binary cross-entropy
\end{itemize}

\noindent The FC-GQCNN model we evaluated against was a pre-trained model from \url{https://berkeleyautomation.github.io/gqcnn/}.

\end{appendices}


\end{document}